\documentclass{article} % For LaTeX2e
\usepackage{iclr2025_conference,times}

\usepackage{amsmath,amsfonts,bm}

\def\eqref#1{equation~\ref{#1}}
\def\1{\bm{1}}

\DeclareMathAlphabet{\mathsfit}{\encodingdefault}{\sfdefault}{m}{sl}
\SetMathAlphabet{\mathsfit}{bold}{\encodingdefault}{\sfdefault}{bx}{n}

\usepackage[hidelinks]{hyperref}
\usepackage{url}
\usepackage{graphicx}
\usepackage{amssymb}
\usepackage{bbm}
\usepackage{subcaption}
\usepackage[capitalise]{cleveref}
\usepackage{enumitem}
\usepackage{booktabs}

\DeclareRobustCommand{\revision}[1]{#1}

\title{Efficient Dictionary Learning with Switch Sparse Autoencoders}

\author{Anish Mudide \thanks{ Correspondence to amudide@mit.edu.} \\
Massachusetts Institute of Technology\\
\And
Joshua Engels \\
Massachusetts Institute of Technology\\
\And
Eric J. Michaud \\
Massachusetts Institute of Technology\\
\And
Max Tegmark \\
Massachusetts Institute of Technology\\
\AND
Christian Schroeder de Witt\\
University of Oxford\\
}

\iclrfinalcopy % Uncomment for camera-ready version, but NOT for submission.
\fancypagestyle{plain}{%
 \fancyhf{} % Clear all header and footer fields
}
\renewcommand{\headrulewidth}{0pt} % Remove header line
\renewcommand{\footrulewidth}{0pt} % Remove footer line
\begin{document}

\maketitle

\begin{abstract}

% ERIC PROPOSED REWRITE

% Thursday morning note from Eric: I don't like my abstract anymore

Sparse autoencoders (SAEs) are a recent technique for decomposing neural network activations into human-interpretable features. However, in order for SAEs to identify all features represented in frontier models, it will be necessary to scale them up to very high width, posing a computational challenge. 
In this work, we introduce \textbf{Switch Sparse Autoencoders}, a novel SAE architecture aimed at reducing the compute cost of training SAEs.
Inspired by sparse mixture of experts models, Switch SAEs route activation vectors between smaller ``expert'' SAEs, enabling SAEs to efficiently scale to many more features.
We present experiments comparing Switch SAEs with other SAE architectures, and find that Switch SAEs deliver a substantial Pareto improvement in the reconstruction vs. sparsity frontier for a given fixed training compute budget.
We also study the geometry of features across experts, analyze features duplicated across experts, and verify that Switch SAE features are as interpretable as features found by other SAE architectures.

% Sparse autoencoders (SAEs) are a recent technique for decomposing language model activations into human interpretable features. However, the current SAE training paradigm to completely explain model behavior is computationally intractable, as it requires scaling to extremely wide SAEs for a single language model. To combat this problem we introduce the Switch SAE, a novel architecture that builds on the mixture of expert architecture to efficiently scale sparse autoencoders to many more features. We present empirical results on GPT-2 that show that the Switch SAE improves the Pareto frontier of flops vs. loss at a given sparsity level. Additionally, we compare the feature geometry of each expert in the mixture, analyze duplicate features that are learned in multiple experts, and show that the Switch SAE features are as interpretable as traditional SAEs.

% To recover all the relevant features from a superintelligent language model, we will likely need to scale sparse autoencoders (SAEs) to billions of features. Using current architectures, training extremely wide SAEs across multiple layers and sublayers at various sparsity levels is computationally intractable. Conditional computation has been used to scale transformers (Fedus et al.) to trillions of parameters while retaining computational efficiency. We introduce the Switch SAE, a novel architecture that leverages conditional computation to efficiently scale SAEs to many more features.
\end{abstract}

\section{Introduction}

Recently, large language models have achieved impressive performance on many tasks~\citep{brown_language_2020}, but they remain largely inscrutable to humans. Mechanistic interpretability aims to open this metaphorical black box and rigorously explain model computations \citep{olah2020zoom}. Specifically, much work in mechanistic interpretability has focused on understanding \textit{features}, the specific human-interpretable variables a model uses for computation \citep{olah2020zoom, linear_representation_hypothesis, engels2024not}.  

Early mechanistic attempts to understand features focused on neurons, but this work was stymied by the fact that neurons tend to be \textit{polysemantic}: they are frequently activated by several completely different types of inputs, making them hard to interpret \citep{olah2020zoom}. One theory for why neurons are polysemantic is \textit{superposition}, the idea that language models represent many more concepts than they have available dimensions \citep{elhage2022superposition}. To minimize interference, the superposition hypothesis posits that features are encoded as almost orthogonal directions in the model's hidden state space. 

\cite{dictionary_monosemanticity_anthropic} and \cite{other_sae_paper} propose to disentangle these superimposed model representations into monosemantic features by training unsupervised sparse autoencoders \revision{\citep{lee2007sparse, le2013building, konda2014zero}} on intermediate language model activations. The success of this technique has led to an explosion of recent work \citep{templeton2024scaling, gao2024scalingevaluatingsparseautoencoders} that has focused on scaling sparse autoencoders to frontier language models such as Claude 3 Sonnet \citep{claude3modelcard} and GPT-4 \citep{gpt_4_tech_report}. Despite scaling SAEs to 34 million features, \cite{templeton2024scaling} estimate that there likely remains orders of magnitude more features. Furthermore, \cite{gao2024scalingevaluatingsparseautoencoders} train SAEs on a series of language models and find that larger models require more features to achieve the same reconstruction error. As model sizes continue to grow, current training methodologies are set to quickly become computationally intractable.
% training SAEs with billions of features at various layers, sublayers and sparsity levels is 

Each training step of a sparse autoencoder generally consists of six major computations: the encoder forward pass, the encoder gradient, the decoder forward pass, the decoder gradient, the latent gradient and the pre-bias gradient. \cite{gao2024scalingevaluatingsparseautoencoders} introduce kernels that leverage the sparsity of the TopK activation function to dramatically optimize all computations \textit{except} the encoder forward pass, which is not sparse. After implementing these optimizations, \cite{gao2024scalingevaluatingsparseautoencoders} find that the training time is bottlenecked by the dense encoder forward pass and the memory is bottlenecked by storing the latent pre-activations. 

In this work, we introduce the Switch Sparse Autoencoder, which to our knowledge is the first work to solve these dual memory and FLOP bottlenecks. The Switch SAE combines the Switch layer \citep{fedus2022switch} with the TopK SAE \citep{makhzani2013k, gao2024scalingevaluatingsparseautoencoders}. At a high level, the Switch SAE is composed of many small expert SAEs and a trainable routing network that determines which expert SAE will process a given input. We demonstrate that the Switch SAE delivers an improvement in training FLOPs vs.~training loss over existing methods.

\subsection{Contributions}
\begin{enumerate}[align=left,leftmargin=*]
    \item In~\cref{sec:switch}, we describe the Switch Sparse Autoencoder architecture and compare it to existing SAE architectures. We also describe our training methodology, which balances reconstruction and expert utilization.
    \item In~\cref{sec:scaling_laws}, we describe scaling laws for reconstruction MSE with respect to FLOPs and parameters. We show that Switch SAEs achieve a lower MSE compared to TopK SAEs using the same amount of training compute.
    \item In~\cref{sec:sparsity_v_error}, we benchmark Switch SAEs against ReLU, Gated and TopK SAEs on the sparsity-reconstruction Pareto frontier.
    \item In~\cref{sec:feature_geometry}, we study feature duplication in Switch SAEs and visualize the global structure of Switch SAE features using t-SNE.
    \item In~\cref{sec:automated_interp}, we demonstrate that Switch SAE features are as interpretable as TopK SAE features.
\end{enumerate}

\begin{figure}
    \centering
    \includegraphics{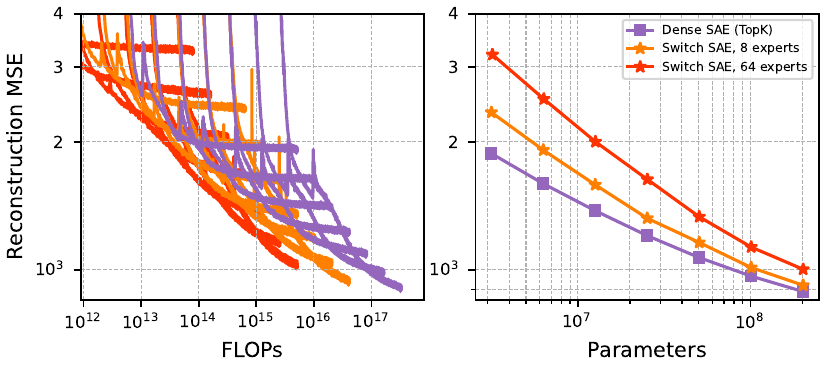}
    \caption{Scaling laws for Switch SAEs. We train dense TopK SAEs and Switch SAEs of varying size with fixed $k=32$. \textbf{Left:} Switch SAEs achieve better reconstruction than dense SAEs at a fixed compute budget. \textbf{Right:} Switch SAEs require more features in total (and therefore more parameters) to achieve the same reconstruction as dense SAEs when trained to convergence, \revision{although this gap narrows for larger SAEs}.}
    \label{fig:scaling-laws}
\end{figure}

\section{Related Work}

\subsection{Mixture of Expert Models}
In a standard deep learning model, every parameter is used for every input. An alternative approach is conditional computation, where only a subset of the parameters are active depending on the input, allowing models with more parameters without the commensurate increase in computational cost. \revision{\cite{jacobs1991adaptive} propose to train multiple networks, where each network is dedicated to a disjoint subset of all possible inputs. Since this seminal work on mixture-of-experts, significant follow-up work has been dedicated to exploring different architectures and configurations \citep{jordan1994hierarchical, tresp2000mixtures, collobert2001parallel, rasmussen2001infinite, aljundi2017expert}.} \cite{shazeer2017outrageously} introduce the Sparsely-Gated Mixture-of-Experts (MoE) layer, the first general purpose conditional computation architecture \revision{component}. A Mixture-of-Experts layer consists of (1) a set of expert networks and (2) a routing network that determines which experts should be active on a given input. 
% The entire model is trained end-to-end, simultaneously updating the routing network and the expert networks. 
 \cite{shazeer2017outrageously} use MoE to scale LSTMs to 137 billion parameters, surpassing the performance of previous dense models on language modeling and machine translation benchmarks. Building on this work, \cite{fedus2022switch} introduce the Switch layer, a simplification to the MoE layer which routes to just a single expert and thereby decreases computational cost and increases training stability. \cite{fedus2022switch} use Switch layers in place of MLP layers to scale transformers to over a trillion parameters. Recent state of the art language models have used MoE layers, including Mixtral 8x7B \citep {jiang2024mixtral} and Grok-1 \citep{grokmodelcard}. To the best of our knowledge, we are the first to apply conditional computation to training SAEs.

\subsection{Deep Learning Training Optimizations}
Our method fits into the wider literature focused on accelerating deep learning training. One type of training speedup uses hardware accelerators like GPUs \citep{raina2009large} and TPUs \citep{jouppi2017datacenter} to optimize highly parallelizable dense matrix operations. Algorithmic improvements, on the other hand, consist of architectural or implementation tricks to speed up forward and backwards passes on fixed hardware. Techniques such as MoE layers (discussed above) and Slide \citep{chen2020slide} utilize sparsity to only evaluate a subset of the parameters for a given wide MLP layer. Other techniques such as Flash Attention \citep{dao2022flashattention} and Reformers \citep{kitaev2020reformer} use hashing or structured matrices to reduce the time complexity of a transformer's attention mechanism. See Appendix A in \cite{dao2022flashattention} for a comprehensive overview of the literature on algorithmic training optimizations. Our work differs from these papers in that we are concerned with not only whether the training optimization results in a speedup, but also whether SAE quality is preserved.

\subsection{Sparse Autoencoders and Improvements}
\revision{Sparse dictionary learning is the task of decomposing input data into a sparse linear combination of basic elements called \textit{atoms}, which together form a \textit{dictionary} \citep{olshausen1997sparse, elad2010sparse}. There exist a wide variety of techniques for performing dictionary learning, such as the method of optimal directions \citep{engan1999method} and K-SVD \citep{aharon2006k}. However, such methods are not scalable to large language models and may not be faithful to the models internals. As such, recent work has focused on applying sparse autoencoders \citep{lee2007sparse, le2013building, konda2014zero}, a simple approximation of sparse dictionary learning, to language models.}

Since the initial works proposing SAEs to separate model representations \citep{other_sae_paper, dictionary_monosemanticity_anthropic}, there have been many proposed improvements to the SAE architecture. These have included alternative activation functions like ProLu~\citep{taggart2024prolu}, TopK\revision{~\citep{makhzani2013k, gao2024scalingevaluatingsparseautoencoders}}, and Batch-TopK~\citep{bussmann2024batchtopk}, architectural changes to solve feature suppression caused by the L1 penalty~\citep{rajamanoharan2024improving, rajamanoharan2024jumping}, and changes to the optimization objective itself~\citep{braun2024identifying}. There are many metrics along which these papers evaluate their improvements, but the most common metric is a Pareto frontier of SAE latent sparsity (measured as average L0), mean squared error, and feature interpretability; thus, these are the metrics we focus on in this paper. We also compare primarily against TopK SAEs~\citep{gao2024scalingevaluatingsparseautoencoders} as our baseline, as recent work \citep{anthropic_august_update} has shown that these achieve state of the art results on these metrics.

\section{The Switch Sparse Autoencoder}
\label{sec:switch}

\subsection{Baseline Sparse Autoencoder}
\label{sec:baseline}

Sparse autoencoders are trained to reconstruct language model activations $\mathbf{x} \in \mathbb{R}^d$ by decomposing them into sparse linear combinations of $M \gg d$ unit-length feature vectors $\mathbf{f}_1, \mathbf{f}_2, \ldots, \mathbf{f}_M \in \mathbb{R}^d$. SAE architectures consist of an encoder $\mathbf{W}_{\text{enc}} \in \mathbb{R}^{M \times d}$, a decoder $\mathbf{W}_{\text{dec}} \in \mathbb{R}^{d \times M}$, bias terms (e.g., $\mathbf{b}_{\text{pre}} \in \mathbb{R}^d$) and an activation function. The TopK SAE~\citep{gao2024scalingevaluatingsparseautoencoders} outputs a reconstruction $\hat{\mathbf{x}}$ of the input activation $\mathbf{x}$, given by
\begin{eqnarray}
\mathbf{z}&=& \text{TopK}(\mathbf{W}_{\text{enc}} (\mathbf{x} - \mathbf{b}_{\text{pre}}))\\
\hat{\mathbf{x}} &=& \mathbf{W}_{\text{dec}} \mathbf{z} + \mathbf{b}_{\text{pre}}
\end{eqnarray}
The latent vector \( \mathbf{z} \in \mathbb{R}^M \) represents how strongly each feature is activated. Since \( \mathbf{z} \) is sparse, the decoder forward pass can be optimized by a suitable kernel. The loss is the reconstruction error $\mathcal{L} = \|\mathbf{x} - \hat{\mathbf{x}}\|_2^2$.

We additionally benchmark against the ReLU SAE~\citep{anthropic_april_update}  and the Gated SAE~\citep{rajamanoharan2024improving}. The ReLU SAE uses the ReLU activation function and applies an L1 penalty to the feature activations to encourage sparsity. The Gated SAE avoids activation shrinkage~\citep{wright2024addressing} by separately determining which features should be active and how strongly activated they should be.

\subsection{Switch Sparse Autoencoder Architecture}
\label{sec:switch_architecture}

\begin{figure}[h]
\begin{center}
\includegraphics[width=5.5in]{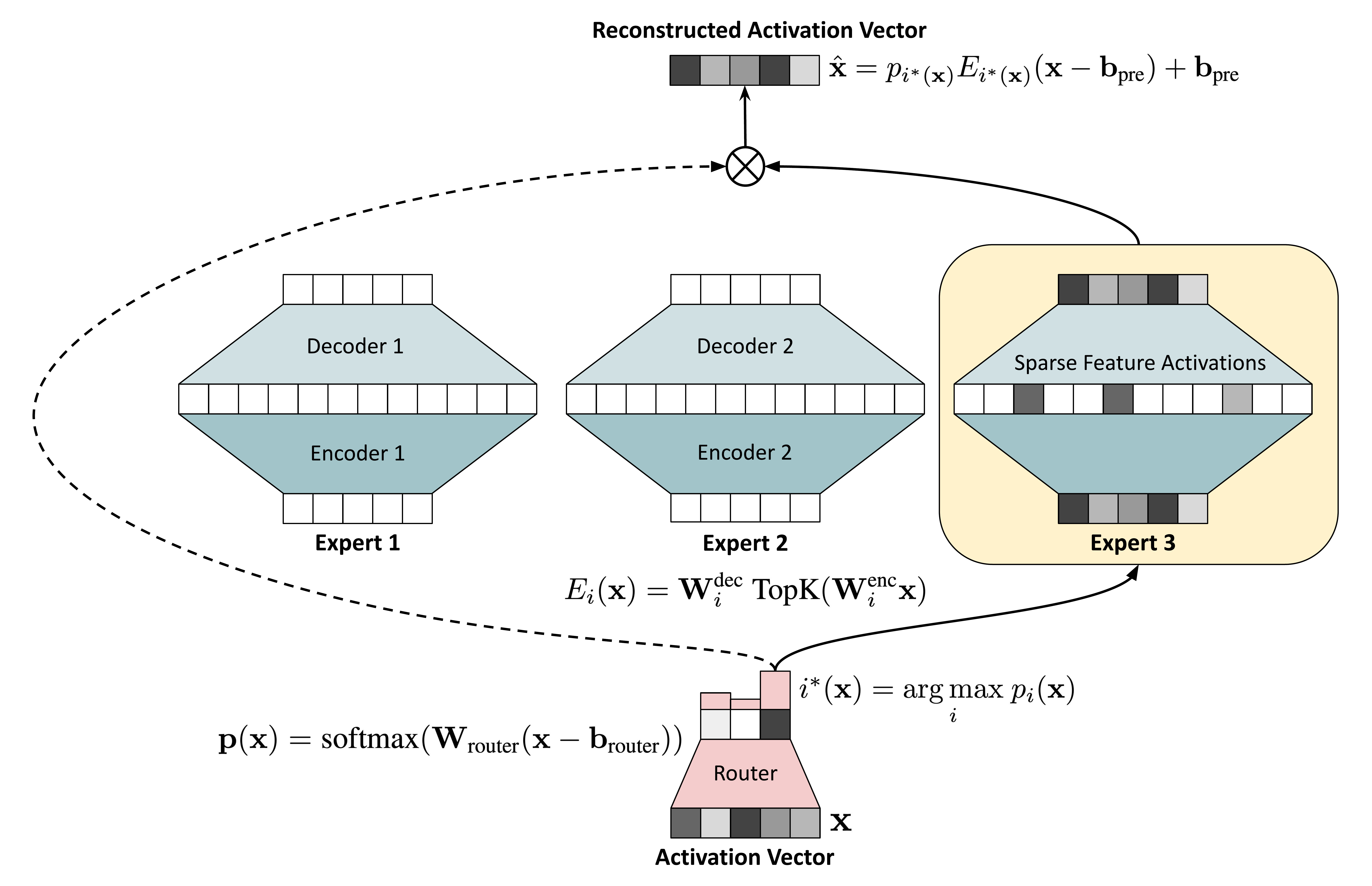}
\end{center}
\caption{Switch Sparse Autoencoder Architecture. The router computes a probability distribution over the expert SAEs and routes the input activation vector to the expert with the highest probability. The figure depicts the architecture for $d=5$, $N=3$, $M=12$.}
\label{fig:architecture}
\end{figure}

The Switch SAE consists of $N$ expert SAEs $\{E_i\}_{i=1}^{N}$ as well as a routing network that determines which expert SAE should be used for a given input (\cref{fig:architecture}). Each expert SAE $E_i$ resembles a TopK SAE with no bias term:
\begin{equation}
E_i(\mathbf{x}) = \mathbf{W}_i^{\text{dec}} \text{TopK}(\mathbf{W}_i^{\text{enc}} \mathbf{x})
\end{equation}
The router, defined by trainable parameters $\mathbf{W}_{\text{router}} \in \mathbb{R}^{N \times d}$ and $\mathbf{b}_{\text{router}} \in \mathbb{R}^d$, computes a probability distribution $\mathbf{p}(\mathbf{x}) \in \mathbb{R}^N$ over the $N$ experts given by $\mathbf{p}(\mathbf{x}) = \text{softmax}(\mathbf{W}_{\text{router}} (\mathbf{x} - \mathbf{b}_{\text{router}}))$. We route the input $\mathbf{x}$ to the most probable expert $i^*(\mathbf{x}) = \underset{i}{\arg\max} \ p_i(\mathbf{x})$. The output $\hat{\mathbf{x}}$ is given by,
\begin{equation}
\hat{\mathbf{x}} = p_{i^*(\mathbf{x})} E_{i^*(\mathbf{x})} (\mathbf{x} - \mathbf{b}_{\text{pre}}) + \mathbf{b}_{\text{pre}}.
\end{equation}
The Switch SAE thus avoids the dense $\mathbf{W}_{\text{enc}}$ matrix multiplication by leveraging conditional computation. When comparing Switch SAEs to existing SAE architectures, we evaluate two cases: (1) FLOP-matched Switch SAEs, which perform the same number of FLOPs per activation but increase the number of features by a factor of $N$, and (2) width-matched Switch SAEs, which reduce the FLOPs per activation by a factor of $N$ while keeping the number of features constant. 
%The Switch SAE avoids the dense $\mathbf{W}_{\text{enc}}$ matrix multiplication. Instead of being one large sparse autoencoder, the Switch Sparse Autoencoder is composed of \( N \) smaller expert SAEs \( \{E_i\}_{i=1}^{N} \). Each expert SAE \( E_i \) resembles a TopK SAE with no bias term:
%\begin{equation}
%E_i(\mathbf{x}) = \mathbf{W}_i^{\text{dec}} \ \text{TopK}(\mathbf{W}_i^{\text{enc}} \mathbf{x})
%\end{equation}
%We route the input to the expert with the highest probability and weight the output by that probability to allow gradients to propagate. We subtract a bias before passing \( \mathbf{x} \) to the selected expert ($i^* = \arg \max_i p_i$) and add it back after weighting by the corresponding probability:
%In total, the Switch Sparse Autoencoder contains \( 2Md + Nd + 2d \) parameters, whereas the TopK SAE has \( 2Md + d \) parameters. The additional \( Nd + d \) parameters we introduce through the router are an insignificant proportion of the total parameters because \( M \gg N \).
%During the forward pass of a TopK SAE, \( Md \) parameters are used during the encoder forward pass, \( kd \) parameters are used during the decoder forward pass and \( d \) parameters are used for the bias, for a total of \( Md + kd + d \) parameters used. Since \( M \gg k \), the number of parameters used is dominated by \( Md \). During the forward pass of a Switch SAE, \( Nd \) parameters are used for the router, \( \frac{M}{N} d \) parameters are used during the encoder forward pass, \( kd \) parameters are used during the decoder forward pass and \( 2d \) parameters are used for the biases, for a total of \( \frac{M}{N} d + kd + Nd + 2d \) parameters used. Since the encoder forward pass takes up the majority of the compute, we effectively reduce the compute by a factor of \( N \). This approximation becomes better as we scale \( M \), which will be required to capture all the safety-relevant features of future language models. Furthermore, the TopK SAE must compute and store \( M \) pre-activations. Due to the sparse router, the Switch SAE only needs to store \( \frac{M}{N} \) pre-activations, improving memory efficiency by a factor of \( N \) as well.
\subsection{Switch Sparse Autoencoder Training}
\label{sec:switch_training}

We train the Switch SAE end-to-end. When computing $\hat{\mathbf{x}}$, we weight $E_{i^*(\mathbf{x})} (\mathbf{x} - \mathbf{b}_{\text{pre}})$ by $p_{i^*(\mathbf{x})}$ to make the router differentiable. We adopt many of the training strategies described in~\cite{templeton2024scaling} and~\cite{gao2024scalingevaluatingsparseautoencoders} (see~\cref{sec:training_details} for details).\footnote{Our code can be found at \url{https://github.com/amudide/switch_sae}}

The TopK SAE enforces sparsity via its activation function and thus directly optimizes the reconstruction MSE. Following~\cite{fedus2022switch}, we train our Switch SAEs using a weighted combination of the reconstruction MSE and an auxiliary loss which encourages the router to send an equal number of activations to each expert to reduce overhead. For a batch $\mathcal{B}$ with $T$ activations, the auxiliary loss is given by
$\mathcal{L}_{\text{aux}} = N \cdot \sum_{i=1}^N f_i \cdot P_i$. $f_i$ represents the proportion of activations that are routed to expert $i$, and $P_i$ represents the proportion of router probability that is assigned to expert $i$. Formally,
$f_i = \frac{1}{T} \sum_{\mathbf{x} \in \mathcal{B}} \mathbbm{1}{\{i^*(\mathbf{x}) = i\}}$ and $P_i = \frac{1}{T} \sum_{\mathbf{x} \in \mathcal{B}} p_i(\mathbf{x})$. The auxiliary loss is minimized when the batch of activations is routed uniformly across the $N$ experts. The reconstruction loss is defined to be $
\mathcal{L}_{\text{recon}} = \frac{1}{T} \sum_{\mathbf{x} \in \mathcal{B}} \|\mathbf{x} - \hat{\mathbf{x}}\|_2^2$. Note that $\mathcal{L}_{\text{recon}} \propto d$. Let $\alpha$ represent a tunable load balancing hyperparameter that scales the auxilliary loss. The total loss $\mathcal{L}_{\text{total}}$ is given by 
$\mathcal{L}_{\text{total}} = \mathcal{L}_{\text{recon}} + \alpha \cdot d \cdot \mathcal{L}_{\text{aux}}$. We optimize $\mathcal{L}_{\text{total}}$ using Adam~\citep{kingma2014adam}. \revision{We find that results are not overly sensitive to the value of $\alpha$, but that $\alpha = 3$ is a reasonable default based on a hyperparameter sweep (see~\cref{sec:hyperparameter} for details).}

\section{Results}
\begin{figure}
    \centering
    \includegraphics{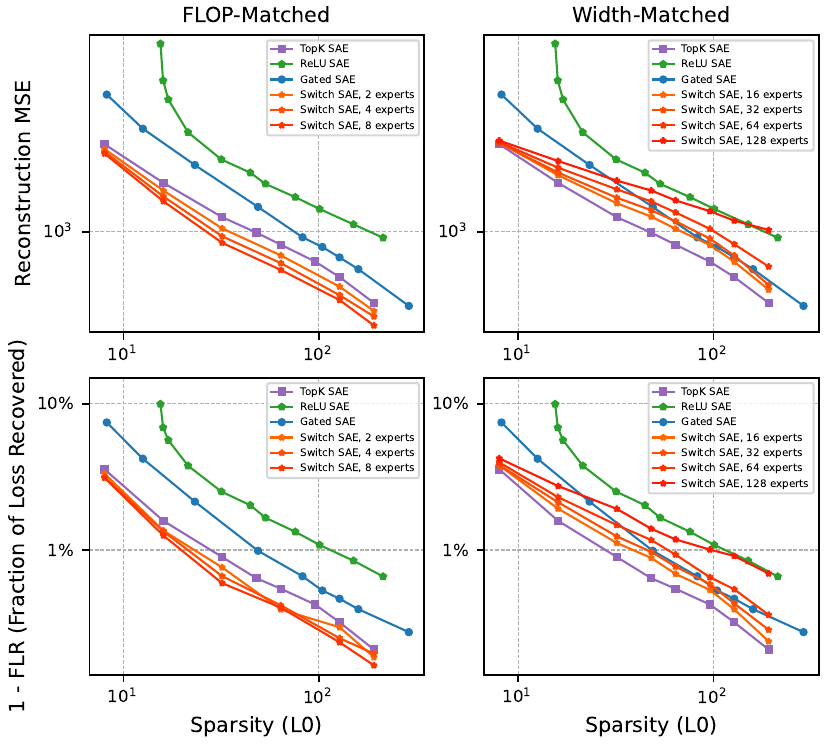}
    \caption{Pareto frontier of sparsity versus (top) reconstruction mean squared error and (bottom) \revision{1 - FLR [fraction of loss recovered]}. FLOP-matched Switch SAEs Pareto-dominate TopK SAEs using the same amount of compute (left). Width-matched Switch SAEs perform slightly worse than TopK SAEs but Pareto-dominate ReLU SAEs while performing fewer FLOPs (right).}
    \label{fig:l0-mse}
\end{figure}
We train sparse autoencoders on the residual stream activations of GPT-2 small, which have a dimension of $768$~\citep{radford2019language}. Early layers of language models handle detokenization and feature engineering, while later layers refine representations for next-token prediction~\citep{lad2024remarkable}. In this work, we follow~\cite{gao2024scalingevaluatingsparseautoencoders} and train SAEs on activations from layer 8, which we expect to hold rich feature representations not yet specialized for next-token prediction. Using text data from OpenWebText~\citep{Gokaslan2019OpenWeb}, we train for 100K steps using a batch size of 8192, for a total of approximately 820 million tokens. 

\subsection{Scaling laws for reconstruction error}
\label{sec:scaling_laws}

We first study scaling laws for Switch SAEs, comparing them to dense TopK SAEs at fixed sparsity $k=32$. In \cref{fig:scaling-laws}, we show scaling curves of reconstruction MSE error with respect to both FLOPs and number of parameters for Switch SAEs with 8 and 64 experts. We find that Switch SAEs have favorable scaling with respect to FLOPs compared to dense TopK SAEs. In fact, Switch SAEs using $\sim$1 OOM less compute can often achieve the same reconstruction MSE as TopK SAEs. However, Switch SAEs perform worse at fixed width relative to dense TopK SAEs. Increasing the number of experts improves compute efficiency but reduces parameter efficiency. We hypothesize that this trade-off is a result of features needing to be duplicated across multiple Switch SAE experts, which we discuss in more detail later. \revision{Nevertheless, in the right subplot of \cref{fig:scaling-laws}, we show that the gap between TopK and Switch SAE performance at a fixed width \textit{shrinks} as we scale the number of parameters. Thus, for large-scale experiments, this gap may be imperceptible; since this is the regime in which efficient training is most useful, we believe that this is not a significant weakness of Switch SAEs.}

\subsection{Sparsity vs. reconstruction error}
\label{sec:sparsity_v_error}

We now study Switch SAE performance in the reconstruction error vs. sparsity frontier. We benchmark the Switch SAE against the ReLU SAE~\citep{dictionary_monosemanticity_anthropic}, the Gated SAE~\citep{rajamanoharan2024improving} and the TopK SAE~\citep{gao2024scalingevaluatingsparseautoencoders}. We present results for two settings:

\begin{enumerate}
    \item FLOP-matched: The ReLU, Gated and TopK SAEs are trained with $32 \cdot 768 = 24576$ features. We train Switch SAEs with 2, 4 and 8 experts. Each expert of the Switch SAE with $N$ experts has 24576 features, for a total of $24576 \cdot N$ features. The Switch SAE performs roughly the same number of FLOPs per activation compared to the TopK SAE.
    \item Width-matched: Each SAE is trained with $32 \cdot 768 = 24576$ features. We train Switch SAEs with 16, 32, 64 and 128 experts. Each expert of the Switch SAE with $N$ experts has $\frac{24576}{N}$ features. The Switch SAE performs roughly $N$ times fewer FLOPs per activation compared to the TopK SAE.
\end{enumerate}

\revision{Note that the router parameters make up a small, insignificant proportion of the total parameters. Across all our experiments, the router parameters make up between $0.002\%$ and $0.3\%$ of the total parameters.} For a wide range of sparsity values \revision{(L0 between 8 and 128)}, we report the reconstruction MSE and the \revision{fraction} of cross-entropy loss recovered \revision{(FLR)} when the sparse autoencoder output is patched into the language model. A \revision{FLR} value of 1 corresponds to a perfect reconstruction, while a \revision{FLR} value of 0 corresponds to a zero-ablation \revision{(setting the residual stream to the zero vector)}.

\subsubsection{FLOP-Matched Results}

For the FLOP-matched experiments, we train Switch SAEs with 2, 4 and 8 experts, with the results shown in the left two plots of \cref{fig:l0-mse}. The Switch SAEs are a Pareto improvement over the TopK, Gated and ReLU SAEs in terms of both MSE and \revision{FLR}. As we scale up the number of experts and represent more features, performance continues to increase while keeping computational costs and memory costs (from storing the pre-activations) roughly constant. \revision{For a detailed discussion of FLOP calculations, see~\cref{sec:flop}.}

\subsubsection{Width-Matched Results}

For the width-matched experiments, we train Switch SAEs with 16, 32, 64 and 128 experts, with the results shown in the right two plots of \cref{fig:l0-mse}. The Switch SAEs consistently underperform compared to the TopK SAE in terms of MSE and \revision{FLR}, while for the most part outperforming Gated and ReLU SAEs. When L0 is low, Switch SAEs perform particularly well. This suggests that the high frequency features that improve reconstruction fidelity the most for a given activation might lie within the same cluster.

% The Switch SAE with 16 experts is a Pareto improvement compared to the Gated SAE in terms of both MSE and loss recovered, despite performing roughly 16x fewer FLOPs per activation. The Switch SAE with 32 experts is a Pareto improvement compared to the Gated SAE in terms of loss recovered. The Switch SAE with 64 experts is a Pareto improvement compared to the ReLU SAE in terms of both MSE and loss recovered. The Switch SAE with 128 experts is a Pareto improvement compared to the ReLU SAE in terms of loss recovered. The Switch SAE with 128 experts is a Pareto improvement compared to the ReLU SAE in terms of MSE, excluding when $k = 192$. The $k = 192$ scenario for the 128 expert Switch SAE is an extreme case: each expert SAE has $\frac{24576}{128} = 192$ features, meaning that the TopK activation is effectively irrelevant. 

%\begin{figure}[h]
%\begin{center}
%\includegraphics[width=4in]{fig/l0_mse.png}
%\end{center}
%\caption{\textbf{L0 vs. MSE for fixed width SAEs.} The 16 expert Switch SAE outperforms the Gated SAE. The 32 and 64 expert Switch SAEs outperform the ReLU SAE. The 128 expert Switch SAE outperforms the ReLU SAE excluding the extreme $k = 192$ setting.}
%\end{figure}

%\begin{figure}[h]
%\begin{center}
%\includegraphics[width=4in]{fig/l0_lossrec.png}
%\end{center}
%\caption{\textbf{L0 vs. Loss Recovered for fixed width SAEs.} The 16 and 32 expert Switch SAEs outperform the Gated SAE. The 64 and 128 expert Switch SAEs outperform the ReLU SAE.}
%\end{figure}

These results demonstrate that Switch SAEs can reduce the number of FLOPs per activation by up to 128x while still retaining the performance of a ReLU SAE. We further believe that Switch SAEs can likely achieve greater acceleration on larger language models.

\revision{To show the generality of Switch SAE training, in \cref{sec:additional_training} we train on four layers of GPT-2 at residual, attention, and MLP outputs. We also train on a single layer of Gemma 2 2B~\citep{team2024gemma}. We find that Switch SAEs perform well at residual layers on all positions and models tested, but worse on MLP and attention outputs.}

%\begin{figure}[h]
%\begin{center}
%\includegraphics[width=4in]{fig/flopmatch_l0_mse.png}
%end{center}
%caption{\textbf{L0 vs. MSE for FLOP-matched SAEs.} The Switch SAEs consistently outperform the TopK, Gated and ReLU SAEs. Performance improves with a greater number of experts.}
%\end{figure}

%\begin{figure}[h]
%\begin{center}
%includegraphics[width=4in]{fig/flopmatch_l0_lossrec.png}
%\end{center}
%\caption{\textbf{L0 vs. Loss Recovered for FLOP-matched SAEs.} The Switch SAEs consistently outperform the TopK, Gated and ReLU SAEs. Performance improves with a greater number of experts.}
%\end{figure}

%Fedus et al. find that their sparsely-activated Switch Transformer is significantly more sample-efficient compared to FLOP-matched, dense transformer variants. We similarly find that our Switch SAEs are 5x more sample-efficient compared to the FLOP-matched, TopK SAE baseline. Our Switch SAEs achieve the reconstruction MSE of a TopK SAE trained for 100K steps in less than 20K steps. This result is consistent across 2, 4 and 8 expert Switch SAEs.

% \begin{figure}[h]
% \begin{center}
% \includegraphics[width=4in]{fig/efficiency.png}
% \end{center}
% \caption{\textbf{Sample efficiency of Switch SAEs compared to the TopK SAE.} Switch SAEs achieve the same MSE as the TopK SAE in 5x fewer training steps.}
% \end{figure}

% Switch SAEs speed up training while capturing more features and keeping the number of FLOPs per activation fixed. Kaplan et al. similarly find that larger models are more sample efficient.

\begin{figure}[t]
    \centering
    \begin{subfigure}[t]{0.31\textwidth}
        \includegraphics[width=\textwidth]{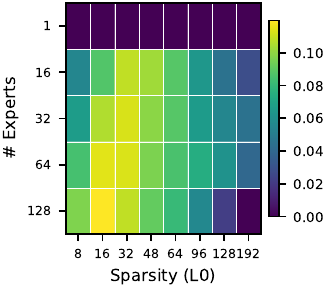}
        \caption{Fraction of SAE decoder vectors with nearest neighbor cosine sim $>0.9$, width-matched SAEs.}
        \label{fig:geo1}
    \end{subfigure}
    \hfill
    \begin{subfigure}[t]{0.31\textwidth}
        \includegraphics[width=\textwidth]{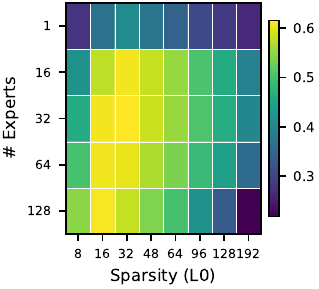}
        \caption{Average cosine sim with nearest neighbor for decoder vectors, width-matched SAEs.}
        \label{fig:geo2}
    \end{subfigure}
    \hfill
    \begin{subfigure}[t]{0.34\textwidth}
        \includegraphics[width=\textwidth]{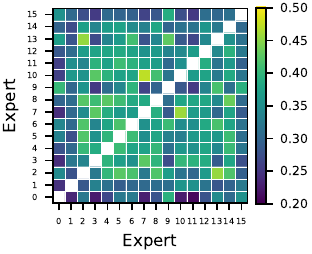}
        \caption{Average cosine sim between nearest neighbors across experts (16 experts, $\text{L0}=64$).}
        \label{fig:geo3}
    \end{subfigure}
    \caption{\revision{Switch SAE feature geometry experiments, measured via cosine similarity between SAE decoder vectors. We find that Switch SAEs with more experts exhibit more feature duplication, but that this effect diminishes for larger L0s. Furthermore, between-expert similarities show that experts specialize moderately; expert 0, for example, has low similarity with most other experts.}}
    \label{fig:all-geo}
\end{figure}

\subsection{Feature Geometry}
\label{sec:feature_geometry}

\subsubsection{Feature Similarity}
We are interested in why Switch SAEs achieve a worse reconstruction MSE than TopK SAEs of the same size. One hypothesis is that because on any given forward pass only a single expert is activated, some experts will have to learn duplicate features, reducing SAE capacity (this is necessary since there is no perfect split of features into clusters such that features in different clusters never co-occur). We test this hypothesis using a common (see e.g. \cite{braun2024identifying}) SAE evaluation metric: nearest neighbor cosine similarity. 

One way to use this metric to measure the number of duplicate features in an SAE is to measure the proportion of vectors in the decoder that have a nearest neighbor above a given cosine similarity, since highly similar decoder vectors are likely duplicates. In \cref{fig:geo1}, we compare SAEs trained with different numbers of experts and different sparsities on this measure with a threshold of $0.9$, and find that as soon as we train with experts this measure jumpy sharply from a baseline of $0$ in TopK SAEs to $5$ to $10$ percent. In other words, strict duplicates likely reduce Switch SAE capacity by up to $10\%$. This effect is less prevalent at larger L0s. However, we are not just worried about exact duplicates, but also a general shift towards redundancy: in \cref{fig:geo2}, we find that a more global metric, cosine similarity of the nearest neighbor averaged across all features, shows a similar pattern across sparsity and number of experts.

Another measure of Switch SAE quality is \textit{expert specialization}, that is, how similar are the sets of features each expert learns. We quantify by averaging the cosine similarity of each feature in expert $A$ with its nearest neighbor in expert $B$. On one end, if the features are perfectly clustered, then each pair of experts should have the same cosine similarity as random blocks of the same size in a TopK SAE of the same size. On the other end, no specialization should result in identical experts. In \cref{fig:geo3}, we plot this metric for all pairs of experts in a $16$ expert Switch SAE with $\text{L0} = 64$. Some experts, e.g. expert $0$, seem unique, while other pairs of experts, e.g. $10$ and $7$, are highly similar. All pairs are more similar than random blocks of the same size in a TopK SAE, which has a value of about $0.2$ under this same metric.

% \subsubsection{Duplicate Features}

% \subsubsection{Cosine Sims}

\subsubsection{T-SNE Visualization}

\begin{figure}
    \centering
    \includegraphics{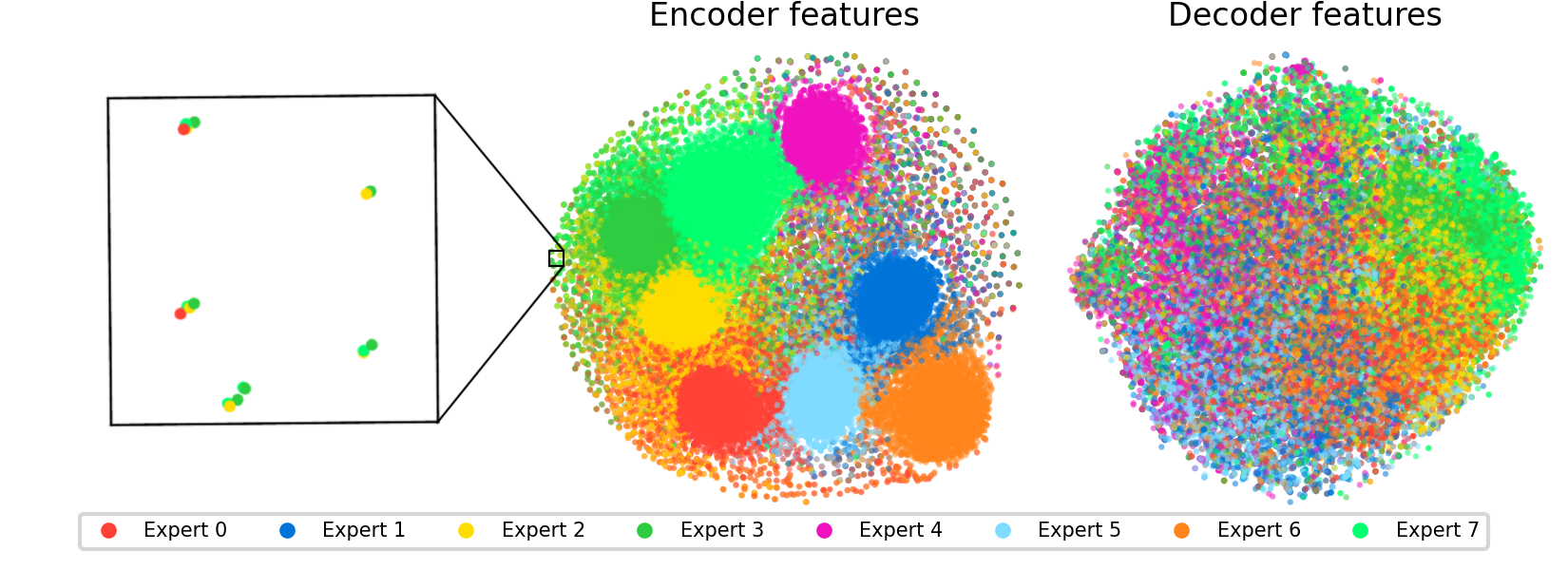}
    \caption{t-SNE projections of encoder and decoder features for a Switch SAE with 65k total features and 8 experts. Encoder features appear to cluster together by expert. Away from these clusters, we see a variety of isolated points which are in fact tight groups of features which have been duplicated across multiple experts. We do not observe as clear clusters in the decoder features.}
    \label{fig:tsne}
\end{figure}

To visualize the global structure of SAE features, we show t-SNE projections of the encoder and decoder feature vectors in \cref{fig:tsne}. We find that encoder features from the same expert cluster together, while decoder features tend to be more diffuse. In the encoder feature t-SNE projection, we can also directly observe feature duplication -- around the periphery of the plot we find a variety of isolated points which upon closer inspection are actually tight groupings of multiple features from different experts.

\subsection{Automated Interpretability}
\label{sec:automated_interp}

\begin{figure}[h]
\begin{center}
\includegraphics[width=\textwidth]{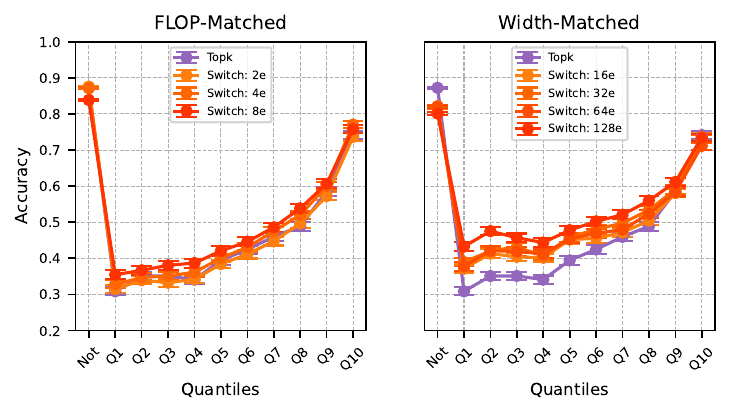}
\end{center}
\caption{Automated interpretability detection results across SAE feature activation quantiles for $1000$ random features, 95\% confidence intervals shown. ``Not'' means that the context comes from a random set of text where the feature was not activating.}
\label{fig:feature_interp}
\end{figure}

\cite{dictionary_monosemanticity_anthropic} evaluate how interpretable sparse autoencoder features using \textit{automated interpretability}: they generate explanations for each feature by giving top activating examples for that feature to a language model, and then ask a language model (in a new context) to predict, using only the description of the feature, on which tokens the described feature fires. More recently, \cite{juang2024autointerp} introduce a more compute efficient method that measures \textit{detection}, whether the explanation allows a language model to predict whether a feature fires at all on a given \textit{context}. \cite{juang2024autointerp} additionally measure detection at each decile of feature activation, as well as on contexts where the feature does not fire at all (where to be correct the model should report that the feature does not fire). Using the \cite{juang2024autointerp} implementation, we find that FLOP-matched SAE features have similar interpretability as TopK SAEs, but width-matched SAEs have a \textit{higher} true positive rate on contexts where the features fire, while at the same time having a lower true negative rate (\cref{fig:feature_interp}). We interpret this result as Switch SAEs being biased towards having duplicate frequent features, which may both be easier to detect and more prone to false positives.

\section{Conclusion}
Switch SAEs are a promising approach towards scaling sparse autoencoders, as they allow an improvement in the FLOPs vs. MSE scaling law without a significant reduction in feature interpretability. We believe that Switch SAEs may find their best application for huge training runs on large clusters of GPUs, since in this setting each expert can be split on to a separate GPU, leading to significant wall clock training speed ups. Overall, we believe that this work provides a case study for applying existing machine learning training optimization techniques to the setting of sparse autoencoders for feature extraction from language models; we hope that the investigations in this paper serve as a  guide for adapting more such techniques.

\textbf{Limitations:} The most significant limitation of the Switch SAE is the reduction in performance of the SAE at a fixed number of parameters\revision{ (especially for attention and MLP layers, see \cref{sec:additional_training})}; future work to bridge this gap might investigate feature deduplication techniques, more sophisticated routing architectures, and multiple active experts. \revision{Additionally, we investigate only the simplest possible mixture of experts architecture; while this allows us to focus on the question of \textit{whether} and \textit{how} mixture of experts training works at all for sparse autoencoders, it leaves open the question of the maximum performance gain mixture of experts style training might allow. Thus, future work could examine more sophisticated mixture of experts architectures like  GShard~\citep{lepikhin2020gshard}, DeepSeekMoE~\citep{dai2024deepseekmoe}, and expert choice routing~\citep{zhou2022mixture} to further optimize performance. Finally, we do not study downstream evaluations of Switch SAEs in this paper; for instance, it is possible that the increased duplication of features between experts complicates feature steering or circuit discovery.}

\subsubsection*{Acknowledgments}
We used the dictionary learning repository~\citep{marks2024dictionary_learning} to train our SAEs. We would like to thank Samuel Marks and Can Rager for advice on how to use the repository. We would also like to thank Jacob Goldman-Wetzler, Achyuta Rajaram, Michael Pearce, Gitanjali Rao, Satvik Golechha, Kola Ayonrinde, Rupali Bhati, Louis Jaburi, Vedang Lad, Adam Karvonen, Shiva Mudide, Sandy Tanwisuth, JP Rivera and Juan Gil for helpful discussions. This work is
supported by Erik Otto, Jaan Tallinn, the Rothberg Family Fund for Cognitive Science, the NSF Graduate Research Fellowship (Grant No. 2141064), IAIFI through NSF
grant PHY-2019786, UKRI grant: Turing AI Fellowship EP/W002981/1, and Armasuisse Science+Technology. AM was
greatly helped by the MATS program, funded by AI Safety Support.

\bibliographystyle{iclr2025_conference}
\bibliography{iclr2025_conference}

\appendix

\section{Implementation Details}
\label{sec:implementation_details}
\subsection{Switch Sparse Autoencoder Training Details}
\label{sec:training_details}

We initialize the rows of $\mathbf{W}_{\text{enc}}^i$ to be parallel to the columns of $\mathbf{W}_{\text{dec}}^i$ for all $i$. We initialize both $\mathbf{b}_{\text{pre}}$ and $\mathbf{b}_{\text{router}}$ to the geometric median of a batch of samples, but we do not tie $\mathbf{b}_{\text{pre}}$ and $\mathbf{b}_{\text{router}}$. We additionally normalize the decoder column vectors to unit-norm at initialization and after each gradient step. We remove gradient information parallel to the decoder feature directions to minimize interference with the Adam optimizer. We set the learning rate based on the $\frac{1}{\sqrt{M}}$ scaling law from~\cite{gao2024scalingevaluatingsparseautoencoders} and linearly decay the learning rate over the last 20\% of training. We do not include neuron resampling~\citep{dictionary_monosemanticity_anthropic} or the AuxK loss~\citep{gao2024scalingevaluatingsparseautoencoders}, but future work could explore employing these strategies to prevent dead latents in larger models. We optimize $\mathcal{L}_{\text{total}}$ with Adam using the default parameters $\beta_1 = 0.9$, $\beta_2 = 0.999$.

\revision{
\subsection{Hyperparameter Search}
\label{sec:hyperparameter}

In our objective function, the balance between reconstruction error and the auxiliary loss (which encourages the router to send an
equal number of activations to each expert) is controlled by the hyperparameter $\alpha$. We train 16-expert and 32-expert Switch SAEs at a fixed sparsity (L0 = 64) and fixed width ($24576$ features), but sweep $\alpha$ across the values $[0.001, 0.003, 0.01, 0.03, 0.1, 0.3, 1, 3, 10, 30, 100]$. We find that the reconstruction MSE is not overly sensitive to the value of $\alpha$ and that $\alpha = 3$ performs the best in both the 16-expert and the 32-expert settings (\cref{fig:alpha_sweep}). We default to $\alpha = 3$ for the rest of our experiments.

\begin{figure}
    \centering
    \includegraphics[width=11cm]{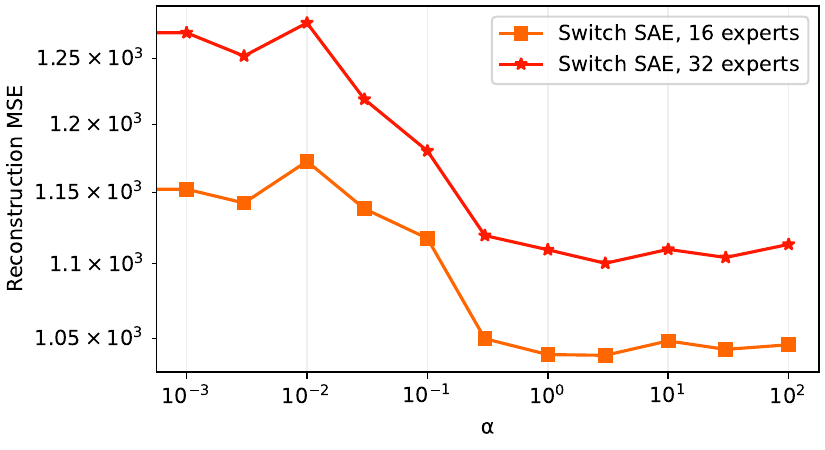}
    \caption{\revision{Hyperparameter sweep for $\alpha$. We train fixed-width Switch SAEs with 16 and 32 experts on activations from GPT-2 small. The sparsity level is set to L0=64.}}
    \label{fig:alpha_sweep}
\end{figure}

\subsection{FLOP Calculations}
\label{sec:flop}

A single training step of a deep neural network requires roughly $6 \cdot P \cdot D$ FLOPs, where $P$ is the number of active parameters and $D$ is the batch size. Let $d$ be the dimension of the language model activations, $M$ be the total number of features and $N$ be the number of experts. For a TopK SAE, $P = 2Md$. For a Switch SAE, $P = dN + \frac{2Md}{N} \approx \frac{2Md}{N}$ since the router parameters are negligible. Thus, we have
\begin{align*}
\text{FLOPS}_{\text{TopK}}(d, M) &= 6 \cdot (2Md) \cdot D, \\
\text{FLOPS}_{\text{Switch}}(d, M, N) &= 6 \cdot \left( dN + \frac{2Md}{N}\right) \cdot D \approx 6 \cdot \left( \frac{2Md}{N}\right) \cdot D.
\end{align*}

}

\revision{
\section{Training on Additional Sites and Models}

\label{sec:additional_training}
In \cref{tab:model_comparison}, we show the results of training on GPT-2 layers $2, 4, 8,$ and $12$ on MLP outputs, attention outputs, and the residual stream. We also train a single SAE on Gemma 2 2B \citep{team2024gemma}. All SAEs are trained with $k = 64$ with a fixed width; we train the Gemma 2 2B SAEs with width $65536$ and the GPT-2 SAEs with width $24576$. We use $8$ experts for the Switch SAEs. As in the main paper, we find that the decrease in SAE performance for a fixed width minimally effects residual stream training; however, Switch SAEs trained on MLP and attention outputs do suffer a significant performance reduction.
}

\begin{table}[h]
\centering
\caption{\revision{Comparison of TopK and Switch SAEs across different models, layers and component types. FVE = Fraction of Variance Explained, FLR = Fraction of Loss Recovered.}}
\begin{tabular}{lcccccccc}
\toprule
Model & Layer & Type & TopK FVE & Switch FVE & TopK FLR & Switch FLR \\
\midrule
GPT-2 & 2 & resid & 0.999 & 0.998 & 0.997 & 0.996 \\
GPT-2 & 2 & attn & 0.919 & 0.884 & 0.997 & 0.968 \\
GPT-2 & 2 & mlp & 0.999 & 0.998 & 0.888 & 0.701 \\
GPT-2 & 4 & resid & 0.997 & 0.997 & 0.996 & 0.995 \\
GPT-2 & 4 & attn & 0.849 & 0.805 & 0.925 & 0.883 \\
GPT-2 & 4 & mlp & 0.877 & 0.819 & 0.853 & 0.777 \\
GPT-2 & 8 & resid & 0.989 & 0.987 & 0.994 & 0.993 \\
GPT-2 & 8 & attn & 0.846 & 0.803 & 0.938 & 0.848 \\
GPT-2 & 8 & mlp & 0.782 & 0.733 & 0.868 & 0.801 \\
GPT-2 & 10 & resid & 0.973 & 0.970 & 0.982 & 0.977 \\
GPT-2 & 10 & attn & 0.892 & 0.863 & 0.942 & 0.883 \\
GPT-2 & 10 & mlp & 0.855 & 0.828 & 0.859 & 0.812 \\
Gemma 2 2B& 12 & resid & 0.961 & 0.955 & 0.991 & 0.989 \\
\bottomrule
\end{tabular}
\label{tab:model_comparison}
\end{table}

%During the forward pass of a TopK SAE, \( Md \) parameters are used during the encoder forward pass, \( kd \) parameters are used during the decoder forward pass and \( d \) parameters are used for the bias, for a total of \( Md + kd + d \) parameters used. Since \( M \gg k \), the number of parameters used is dominated by \( Md \). During the forward pass of a Switch SAE, \( Nd \) parameters are used for the router, \( \frac{M}{N} d \) parameters are used during the encoder forward pass, \( kd \) parameters are used during the decoder forward pass and \( 2d \) parameters are used for the biases, for a total of \( \frac{M}{N} d + kd + Nd + 2d \) parameters used. Since the encoder forward pass takes up the majority of the compute, we effectively reduce the compute by a factor of \( N \). This approximation becomes better as we scale \( M \), which will be required to capture all the safety-relevant features of future language models. Furthermore, the TopK SAE must compute and store \( M \) pre-activations. Due to the sparse router, the Switch SAE only needs to store \( \frac{M}{N} \) pre-activations, improving memory efficiency by a factor of \( N \) as well.

%\section{Spectral Clustering}
%\label{sec:spectral_clustering}

\end{document}